\documentclass{article}
\PassOptionsToPackage{numbers, compress}{natbib}

\usepackage[preprint]{neurips_2024}

\usepackage[utf8]{inputenc} 
\usepackage[T1]{fontenc}    
\usepackage{hyperref}       
\usepackage{url}            
\usepackage{booktabs}       
\usepackage{amsfonts}       
\usepackage{nicefrac}       
\usepackage{microtype}      
\usepackage{xcolor}         
\usepackage{graphicx}
\usepackage{natbib}
\usepackage{amsmath}
\usepackage{cleveref}
\usepackage{geometry}
\usepackage{algorithm}
\usepackage{algorithmic}
\usepackage{multirow} 
\usepackage{wrapfig}
\hypersetup{
hidelinks,
colorlinks=true,
}

\title{Dreamer XL: Towards High-Resolution Text-to-3D Generation via Trajectory Score Matching}

\author{
    \textbf{Xingyu Miao}\textsuperscript{1}\quad 
    \textbf{Haoran Duan}\textsuperscript{2}\quad 
   \textbf{ Varun Ojha}\textsuperscript{2}\quad \\
   \textbf{Jun Song}\textsuperscript{3}\quad 
    \textbf{Tejal Shah}\textsuperscript{2}\quad 
    \textbf{Yang Long}\textsuperscript{1}\quad 
    \textbf{Rajiv Ranjan}\textsuperscript{2}\quad \\
\\
\small
    \textsuperscript{1}Durham University, UK\quad \textsuperscript{2}Newcastle University, UK \quad 
    \textsuperscript{3}China University of Geosciences, China\\
}

\begin{document}

\maketitle

\begin{abstract}
In this work, we propose a novel Trajectory Score Matching (TSM) method that aims to solve the pseudo ground truth inconsistency problem caused by the accumulated error in Interval Score Matching (ISM) when using the Denoising Diffusion Implicit Models (DDIM) inversion process. Unlike ISM which adopts the inversion process of DDIM to calculate on a single path, our TSM method leverages the inversion process of DDIM to generate two paths from the same starting point for calculation. Since both paths start from the same starting point, TSM can reduce the accumulated error compared to ISM, thus alleviating the problem of pseudo ground truth inconsistency. TSM enhances the stability and consistency of the model's generated paths during the distillation process. We demonstrate this experimentally and further show that ISM is a special case of TSM. Furthermore, to optimize the current multi-stage optimization process from high-resolution text to 3D generation, we adopt Stable Diffusion XL for guidance. In response to the issues of abnormal replication and splitting caused by unstable gradients during the 3D Gaussian splatting process when using Stable Diffusion XL, we propose a pixel-by-pixel gradient clipping method. Extensive experiments show that our model significantly surpasses the state-of-the-art models in terms of visual quality and performance. Code: \url{https://github.com/xingy038/Dreamer-XL}.
\end{abstract}

\section{Introduction}
In recent years, Virtual Reality (VR) and Augmented Reality (AR) have increasingly become a part of our daily lives, and the demand for high-quality 3D content has increased significantly. 3D technology has become extremely important, allowing us to visualize, understand and interact with complex objects and environments. It also plays a key role in various fields such as architecture, animation, gaming and virtual reality. In addition, 3D technology shows broad application prospects in retail \cite{wodehouse20163d}, online meetings \cite{nakanishi1999freewalk}, education \cite{reisouglu20173d} and other fields \cite{miao2024conrf, Dynamicduan}. Despite its wide application, the complexity of creating 3D content poses considerable challenges: generating high-quality 3D models requires computional time, effort, and expertise. Given these challenges, methods for generating 3D from text have become particularly important in recent years \cite{ lin2023magic3d, zhu2023hifa, ma2023geodream, shi2023mvdream}. These methods create accurate 3D models directly from natural language descriptions, thereby reducing manual input in traditional 3D modeling processes. Once the text-to-3D method can efficiently generate large amounts of data, it will not only shorten the production time of 3D content, but also reduce costs and improve production efficiency. 

\begin{figure}
    \centering
    \includegraphics[width=\linewidth]{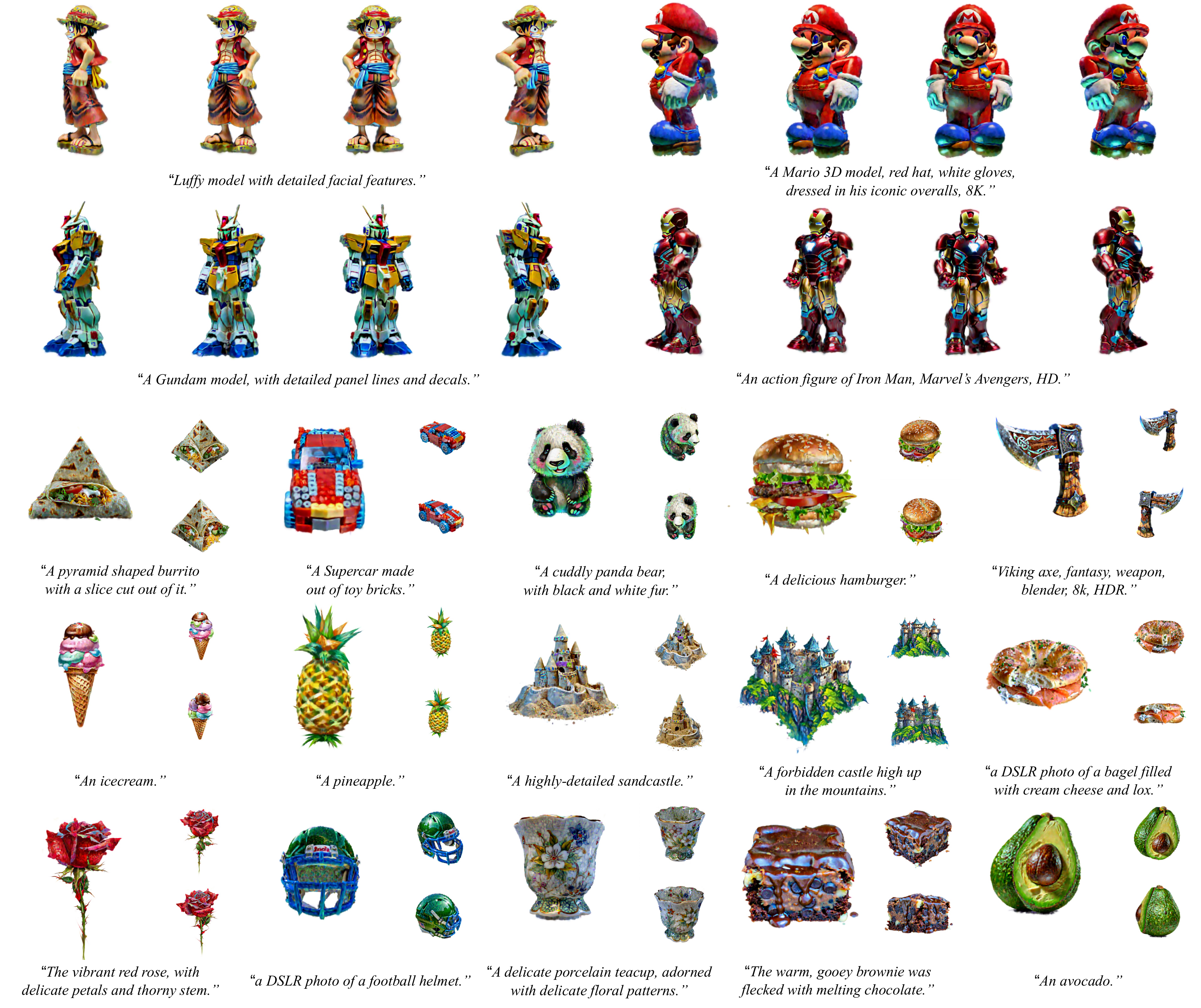}
    \caption{\textbf{Example of text-to-3D content generated from scratch by our Dreamer XL.} Our Dreamer XL is based on 3D Gaussian splatting using stable diffusion XL. Please zoom in for details.}
    \label{fig:teaser}
\end{figure}

Typically, text-to-3D generation methods utilize pre-trained text-to-image diffusion models \cite{saharia2022photorealistic} as an image prior to training neural parametric 3D models such as Neural Radiance Fields (NeRF) \cite{mildenhall2021nerf} and 3D Gaussian splitting \cite{kerbl3Dgaussians}. These approaches enable the rendering of consistent images that are aligned with the text. This process essentially relies on Score Distillation Sampling (SDS) \cite{poole2022dreamfusion}. Through SDS, the model can distill the capabilities of the pre-trained 2D diffusion model to obtain rendered images, and optimize the parameters of the 3D model through backpropagation, so that the 3D model can be effectively trained even without actual image data. However, since random noise generates inconsistent pseudo-baselines, the results obtained by SDS optimization of 3D models tend to be averaged, leading to problems such as over-smoothing. 

Although some recent work \cite{wang2024prolificdreamer, liang2023luciddreamer} devote themselves to solving the over-smoothing problems, they inevitably lead to the generation of low-resolution and average results due to the inherent limitations of stable diffusion models and their sampling methods. For example, \cite{liang2023luciddreamer}, inspired by the DDIM inversion process, proposed interval score matching, which can generate relatively consistent results. However, due to the inherent cumulative errors in the DDIM inversion process, it may lead to the averaging of results in certain regions. Furthermore, most existing methods do not yet support the new high-resolution Stable Diffusion XL (SDXL) \cite{podell2023sdxl, luo2023latent}. To achieve an output of 1024x1024 high resolution, multi-stage optimization is necessary. The main reason is the inherent instability of the Variational Autoencoder (VAE) in the SDXL \cite{podell2023sdxl} architecture, which is particularly evident during the optimization process of the 3D Gaussian Splatting. In this process, the gradients directly affect the duplication and deletion of point clouds in 3D space. The anomalous gradients introduced by SDXL severely hinder the optimization process of 3D Gaussian Splatting, leading to generated 3D models that lose complex texture details, have blurred appearances, and exhibit abnormal colors. In severe cases, this can cause the 3D models to fail to converge. 

In this work, we aim to overcome the above limitations. The reverse process of DDIM was adopted by \cite{liang2023luciddreamer}, effectively reducing the higher reconstruction errors generated by the diffusion model's one-step reconstruction and generating relatively consistent pseudo-ground truth. However, the inherent accumulation errors in the reverse process of DDIM still result in semantic changes in these pseudo-ground truths, leading to partially averaged reconstruction results in certain regions, resulting in erroneous and unrealistic outcomes in these regions. To address this issue, we propose a novel method called Trajectory Score Matching (TSM). We through simple improvements to Interval Score Matching (ISM) \cite{liang2023luciddreamer}, effectively alleviate the average effect of inconsistent pseudo-ground truths caused by inherent accumulation errors. We demonstrate that our TSM has smaller accumulation errors compared to ISM, and ISM can be considered as a special case of TSM. Through experiments, we prove that the effects produced by our TSM are superior to ISM, yielding highly realistic and detailed results. In order to achieve high-resolution output, previous methods need to undergo multi-stage training. Our model directly uses SDXL that supports high resolution as guidance without going through multi-stage training. This not only reduces training costs, but also simplifies the training process. However, we still need to solve the inherent gradient instability problem of SDXL. Therefore, we propose a pixel-by-pixel gradient clipping method, which effectively alleviates the inherent gradient instability of SDXL. In summary, the contributions of our work are as follows:

\begin{itemize}
  \item We investigate and analyze the inherent accumulated errors produced by DDIM inversion process in interval score matching (ISM), resulting in the presence of inconsistent pseudo-ground truth.
  
  \item To address the aforementioned limitation, we introduce a novel Trajectory Score Matching (TSM) method. Unlike the single path of ISM, TSM improves ISM to a dual path, effectively alleviating the inconsistent pseudo-ground truth issue generated from the inherent accumulated error of DDIM.

  \item To simplify the training process and generate high-resolution, high-quality text-to-3D results. We are the first to leverage SDXL for guidance based on 3D Gaussian splatting. In addition, we also introduce a novel gradient clipping method, which effectively solves the problem of SDXL in gradient stability. Extensive experiments demonstrate that our method significantly outperforms the current state-of-the-art methods.
  
\end{itemize}

\section{Related Work}
\paragraph{2D diffusion.}
Score-based generative models and diffusion models \cite{song2019generative, song2020score, ho2020denoising, balaji2022ediff, saharia2022photorealistic, podell2023sdxl} have shown excellent performance in image synthesis \cite{dhariwal2021diffusion}, especially by introducing latent diffusion models (LDM) \cite{rombach2022high} into stable diffusion to generate high-resolution images in latent space. Podel et al. \cite{podell2023sdxl} further extended this model to a larger latent space in SDXL, VAE and U-net, achieving higher resolution (1024×1024). Zhang et al. \cite{zhang2023adding} enhanced the functionality of these models by generating controllable images for different input types. At the same time, the diffusion model also showed impressive performance in converting text into image synthesis, which opens up the possibility of using this technology to directly generate 3D images from text \cite{chen2023fantasia3d, hong2022avatarclip, lin2023magic3d, michel2022text2mesh, poole2022dreamfusion, wang2024prolificdreamer}.

\paragraph{Text-to-3D Generation.}
Early attempts from text-to-3D are mainly guided by the use of multi-modal information from CLIP \cite{radford2021learning} to achieve information conversion from text-to-3D, with DreamField \cite{jain2021dreamfields} being a pioneer in this direction. However, the multi-modal information of CLIP can only provide rough alignment, and the results of using it for 3D distillation are often unsatisfactory. Zero-1-to-3 \cite{liu2023zero1to3} first introduces camera parameters to fine-tune the 2D pre-trained stable diffusion model, enabling it to generate multi-view images, and then use the multi-view images for 3D reconstruction. This improvement not only facilitates more accurate 3D reconstructions, but also inspires a wealth of derivative research \cite{metzer2023latent, liu2024one, qian2023magic123, liu2023one, shi2023zero123++}. In addition, another research direction explores the possibility of using pre-trained 2D diffusion models to directly optimize 3D representations. These methods often combine differentiable 3D representation techniques, such as NeRF \cite{mildenhall2021nerf}, NeuS \cite{wang2021neus}, and 3D Gaussian Splatting \cite{kerbl3Dgaussians}, and optimize model parameters through backpropagation techniques. Dreamfusion \cite{poole2022dreamfusion} first introduces SDS to optimize 3D representations directly from pre-trained 2D text-to-image diffusion models. Similarly, Score Jacobian Chaining \cite{wang2023score} proposes an alternative method that achieves parameterization effects similar to SDS. ProlificDreamer \cite{wang2024prolificdreamer} conducted an in-depth analysis of the objective function of SDS and proposed a particle-based variational framework called Variational Score Distillation (VSD), which significantly improves the quality of generated content. The latest research combines SDS with Gaussian Splatting to accelerate the optimization process. Consistent3D \cite{wu2024consistent3d} analyzes SDS from the latest perspective of ordinary differential equations (ODE) and proposes a method called Consistency Distillation Sampling (CSD) to solve the challenges of SDS in over-smoothing and inconsistency issues. Similarly, LucidDreamer \cite{liang2023luciddreamer} analyzed the loss function of SDS and proposed interval score matching (ISM), which is very similar to the idea of CSD. However, ISM utilizes the reversible diffusion trajectory of DDIM \cite{song2020denoising} when calculating the two interval steps. DDIM will inevitably produce inherent accumulated errors in this process, resulting in inconsistent reconstruction results. In this work, we empirically follow the mature mainstream architecture method of 3D Gaussian Splatting \cite{liang2023luciddreamer, tang2023dreamgaussian, yi2023gaussiandreamer} as the baseline of our approach. On this basis, inspired by the recent Consistency trajectory model \cite{kim2023consistency}, we propose to optimize the 3D model from two trajectories, and use the less noisy trajectory to guide another noisier trajectory to alleviate the inconsistency problem.

\section{Methods}
This section presents the preliminaries on the inverse DDIM process, SDS and ISM (see \Cref{sec:3.1}). We then propose the Trajectory Score Matching (TSM) method (see \Cref{sec:3.2}), which generates dual paths from the same starting point using the reverse process of DDIM. This enhances the stability and consistency of the model along the entire generative path during the distillation process. We further indicate that ISM is a special case of TSM (see \Cref{sec:3.3}). Additionally, we investigate the challenges of optimizing 3D models using the SDXL architecture (see \Cref{sec:3.4}).
\subsection{Preliminaries}
\label{sec:3.1}
\paragraph{Review of DDIM inversion}
We first consider the most common sampling scheme is that of DDIM \cite{song2020denoising} where intermediate steps are calculated as:
\begin{equation}
x_{t-1} = \sqrt{\alpha_{t-1}} \left(\frac{x_t - \sqrt{1 - \alpha_t} \epsilon_{\phi}(x_t, t, \emptyset_{})}{\sqrt{\alpha_t}}\right) + \sqrt{1 - \alpha_{t-1}} \epsilon_{\phi}(x_t, t, \emptyset_{}),
\end{equation}
where $x_t$ and $x_{t-1}$ represent the noisy latent, $\{\alpha_t\}_{t=0}^T$ (where $a_0=1, a_T=0$) indicates a set of time steps indexing a strictly monotonically increasing noise schedule. The $\epsilon_{\phi}(x_t, t, \emptyset_{})$ denotes the predicted denoising direction (by stable diffusion model) with the given condition $y$ (Here, the condition is null, i.e., unconditioned $\emptyset_{}$).

As noted in DDIM \cite{song2020denoising}, the above denoising process can approximate the inverse transition from $x_t$ to $x_{t-1}$, which can be expressed as:
\begin{equation}
\begin{aligned}
\label{DDIM_inversion}
x_{t} &= \sqrt{\alpha_t} \left(\frac{x_{t-1} - \sqrt{1 - \alpha_t} \epsilon_{\phi}(x_t, t, \emptyset_{})}{\sqrt{\alpha_{t-1}}}\right) + \sqrt{1 - \alpha_{t-1}} \epsilon_{\phi}(x_t, t, \emptyset_{}) \\
&\approx \sqrt{\alpha_t} \left(\frac{x_{t-1} - \sqrt{1 - \alpha_t} \epsilon_{\phi}(x_{t-1}, {t-1}, \emptyset_{})}{\sqrt{\alpha_{t-1}}}\right) + \sqrt{1 - \alpha_{t-1}} \epsilon_{\phi}(x_{t-1}, t-1, \emptyset_{}),
\end{aligned}
\end{equation}
where the approximation is a linearization assumption that $\epsilon_{\phi}(x_t, t, \emptyset_{}) \approx \epsilon_{\phi}(x_{t-1}, t-1, \emptyset_{})$. This approximation inevitably introduces errors, resulting in inconsistencies between the diffusion states in the forward and backward processes.

\begin{figure}
    \centering
    \includegraphics[width=\linewidth]{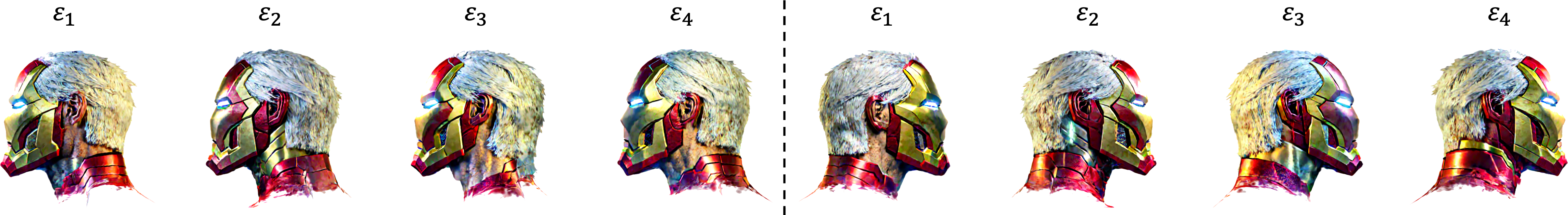}
    \caption{\textbf{ISM example \cite{liang2023luciddreamer}.} We notice that using the same initial value $x_0$ but under different noise $\{\epsilon_1,\epsilon_2,\epsilon_3,\epsilon_4\}$, the generated results still show certain inconsistencies. This is due to the error accumulation inherent in the DDIM inversion process. These inconsistencies can lead to errors or inconsistencies in some areas during the optimization of the 3D model.}
    \label{fig:1}
\end{figure}

\paragraph{Text-to-3D generation by interval score matching (ISM)}
The concepts of the ISM is first introduced by LucidDreamer \cite{liang2023luciddreamer} to address the issues of over-smoothness and inconsistency inherent in the original SDS method. The 3D model leverages a differentiable function $x = g(\theta,c)$ to render images, where $\theta$ represents the trainable 3D parameters and $c$ is camera parameter. The gradient of the ISM loss for $\theta$ is expressed as follows:
\begin{equation}
    \nabla_{\theta} \mathcal{L}_{\text{ISM}}(\theta) := \mathbb{E}_{t,c}\left[\omega(t)(\epsilon_\phi(x_t, t, y) - \epsilon_\phi(x_s, s, \emptyset)) \frac{\partial x}{\partial\theta}\right],
\end{equation}
where $0<s<t$, the noisy latent $x_t$ and $x_s$ are calculated by DDIM inversion process, the $\epsilon_{\phi}(x_t, t, y)$ is the predicted denoising direction given the conditioned $y$, and $\omega(t)$ is a time-dependent weighting function. In DDIM inversion process, the generation of accumulated error is inevitable. Especially for conditional models, these errors are further magnified. Therefore, inconsistent pseudo-ground truths will be generated when optimizing the 3D model, thus affecting the optimization quality of the final result.

\subsection{Trajectory Score Matching}
\label{sec:3.2}
To address the inherent accumulated error in the DDIM inversion process, which leads to the production of inconsistent pseudo-ground truths and consequently suboptimal 3D models. Inspired by recent work \cite{kim2023consistency}, we propose a new approach, Trajectory Score Matching (TSM), which utilizes dual paths originating from the same starting point to minimize error accumulation during iterations. Specifically, similar to ISM \cite{liang2023luciddreamer}, our TSM also utilizes the DDIM inversion to predict an invertible noisy latent trajectory. For a given timestep $s$ (where $0 < s < t$), the corresponding noise latent $x_s$ can be obtained using \Cref{DDIM_inversion}. Considering $x_s$ as the starting latent, it is possible to approximate two noise latents, $x_{\mu}$ and $x_{t}$, on the latent trajectory, where $0 < s < \mu < t \leq T$. This can be expressed as follows:
\begin{equation}
x_\mu = \sqrt{\alpha_\mu} \left(\frac{x_s - \sqrt{1 - \alpha_\mu} \epsilon_{\phi}(x_s, s, \emptyset_{})}{\sqrt{\alpha_s}}\right) + \sqrt{1 - \alpha_s} \epsilon_{\phi}(x_s, s, \emptyset),
\end{equation}
\begin{equation}
x_t = \sqrt{\alpha_t} \left(\frac{x_s - \sqrt{1 - \alpha_t} \epsilon_{\phi}(x_s, s, \emptyset_{})}{\sqrt{\alpha_s}}\right) + \sqrt{1 - \alpha_s} \epsilon_{\phi}(x_s, s, \emptyset).
\end{equation}
Then, we can integrate DDIM inversion and DDIM denoising with the same step size. We define the naive objective of 3D distillation as follows:
\begin{equation}
L_{\text{TSM}}(\theta) := \mathbb{E}_{t,c}\left[\omega(t) ||\epsilon_\phi(x_t, t, y) - \epsilon_\phi(x_\mu, \mu, \emptyset)||^2\right],
\end{equation}
where $x_t$ and $x_\mu$ is generated through DDIM inversion from $x_0$. Followiing \cite{liang2023luciddreamer}, the gradient of TSM loss over $\theta$ is:
\begin{equation}
    \nabla_{\theta} \mathcal{L}_{\text{TSM}}(\theta) := \mathbb{E}_{t,c}\left[\omega(t)(\epsilon_\phi(x_t, t, y) - \epsilon_\phi(x_\mu, \mu, \emptyset)) \frac{\partial x}{\partial\theta}\right].
\end{equation}
The optimization goal of TSM is to maintain the consistency of $x_0$ updates as much as possible to reduce the error introduced by DDIM inversion. Since TSM uses the same noise latent during the inversion process, its cumulative error is relatively small. The algorithm flow of TSM is shown in the \Cref{alg:TSM}. Among them, the blue part marks the differences from ISM.

\begin{algorithm}[!t] 
    \caption{Trajectory Score Matching}
    \label{alg:TSM}
    \begin{algorithmic}[1]
        \STATE Initialization: DDIM inversion step size $\delta_T$ and $\delta_S$, the target prompt $y_{}$, \textcolor{blue}{the offset rate $\gamma \in [0,1]$}
        \WHILE{$\theta$ is not converged}
            \STATE Sample: $x_0 = g(\theta, c), t \sim \mathcal{U}(1, 1000)$
            \STATE let $s = t - \delta_T$, $n = s / \delta_S$, and \textcolor{blue}{$\mu=s+\gamma \delta_T$}            
            \FOR{$i = [0, ..., n-1]$}
                \STATE $x_{(i+1)\delta_S} = \sqrt{\alpha_{(i+1)\delta_S}} \left(\frac{x_{i\delta_S} - \sqrt{1 - \alpha_{(i+1)\delta_S}} \epsilon_{\phi}(x_{i\delta_S}, i\delta_S, \emptyset_{})}{\sqrt{\alpha_{i\delta_S}}}\right) + \sqrt{1 - \alpha_s} \epsilon_{\phi}(x_{i\delta_S}, {i\delta_S}, \emptyset)$
            \ENDFOR
            \STATE predict $\epsilon_\phi(x_s, s, \emptyset)$, then step $x_s \rightarrow x_t$ and \textcolor{blue}{$x_s \rightarrow x_\mu$} via\\
                   $x_t = \sqrt{\alpha_t} \left(\frac{x_s - \sqrt{1 - \alpha_t} \epsilon_{\phi}(x_s, s, \emptyset_{})}{\sqrt{\alpha_s}}\right) + \sqrt{1 - \alpha_s} \epsilon_{\phi}(x_s, s, \emptyset)$\\
                   \textcolor{blue}{$x_\mu = \sqrt{\alpha_\mu} \left(\frac{x_s - \sqrt{1 - \alpha_\mu} \epsilon_{\phi}(x_s, s, \emptyset_{})}{\sqrt{\alpha_s}}\right) + \sqrt{1 - \alpha_s} \epsilon_{\phi}(x_s, s, \emptyset)$}
            \STATE predict $\epsilon_\phi(x_t, t, y)$, \textcolor{blue}{$\epsilon_\phi(x_\mu, \mu, \emptyset)$} and compute TSM gradient \\
                   $\nabla_\theta L_{\text{TSM}} = \omega(t)(\epsilon_\phi(x_{t}, t, y_{}) - \textcolor{blue}{\epsilon_\phi(x_{\mu}, \mu, \emptyset_{})})$

            \STATE update $x_0$ with $\nabla_\theta L_{\text{TSM}}$
        \ENDWHILE
    \end{algorithmic}
\end{algorithm}

\subsection{Comparison with ISM}
\label{sec:3.3}
We now analyze the differences between TSM and ISM theoretically.
\paragraph{TSM has smaller error}
For ISM, the optimized goal is the minimum predicted noise from noise latent $x_s$ and $x_t$. The noise latent can be obtained from \Cref{DDIM_inversion} and then using the pre-trained stable diffusion model to predict noise. During the inversion process of DDIM, $x_s$ is approximated by the prediction noise at $s-1$, and $x_t$ is approximated by the prediction noise at $s$. For simplicity, we can regard the optimization goal of ISM as minimizing noise latent, thus the accumulated error generated by ISM during the optimization process can be expressed as:
\begin{equation}
\label{ISM_error}
\begin{aligned}
x_t - x_s & = \sqrt{\alpha_t} \left(\frac{x_s - \sqrt{1 - \alpha_t} \epsilon_{\phi}(x_s, s, \emptyset_{})}{\sqrt{\alpha_s}}\right) + \sqrt{1 - \alpha_s} \epsilon_{\phi}(x_s, s, \emptyset) \\ &- \sqrt{\alpha_s} \left(\frac{x_{s-1} - \sqrt{1 - \alpha_s} \epsilon_{\phi}(x_{s-1}, {s-1}, \emptyset_{})}{\sqrt{\alpha_{s-1}}}\right) + \sqrt{1 - \alpha_{s-1}} \epsilon_{\phi}(x_{s-1}, {s-1}, \emptyset)
\end{aligned}
\end{equation}
For our TSM, we can also give similar expresses as:
\begin{equation}
\label{TSM_error}
\begin{aligned}
x_t - x_\mu & = \sqrt{\alpha_t} \left(\frac{x_s - \sqrt{1 - \alpha_t} \epsilon_{\phi}(x_s, s, \emptyset_{})}{\sqrt{\alpha_s}}\right) + \sqrt{1 - \alpha_s} \epsilon_{\phi}(x_s, s, \emptyset) \\ &- \sqrt{\alpha_\mu} \left(\frac{x_s - \sqrt{1 - \alpha_\mu} \epsilon_{\phi}(x_s, s, \emptyset_{})}{\sqrt{\alpha_s}}\right) + \sqrt{1 - \alpha_s} \epsilon_{\phi}(x_s, s, \emptyset)
\end{aligned}
\end{equation}
Compared with ISM, which is optimized for a single path, TSM is optimized for a dual path starting from the same noise latent. We show that the accumulated error is smaller in TSM, as shown below.
\paragraph{Theorem 1} \textit{Consider three timesteps $0 < s < \mu < t$, where $\mu$ is defined as $\mu = \gamma(t-s) + s$, with $\gamma\in[0,1]$. Then, we have:}
\begin{equation}
x_t - x_s > x_t - x_\mu \Rightarrow x_\mu > x_s,
\end{equation}

\textit{Proof:} Assume $x_\mu \leq x_s$. Utilizing \Cref{ISM_error} and \Cref{TSM_error}, we aim to demonstrate that:
\begin{equation}
\begin{aligned}
&\sqrt{\alpha_\mu} \left(\frac{x_s - \sqrt{1 - \alpha_\mu} \epsilon_{\phi}(x_s, s, \emptyset_{})}{\sqrt{\alpha_s}}\right) + \sqrt{1 - \alpha_s} \epsilon_{\phi}(x_s, s, \emptyset)\\
&\leq\sqrt{\alpha_s} \left(\frac{x_{s-1} - \sqrt{1 - \alpha_s} \epsilon_{\phi}(x_{s-1}, {s-1}, \emptyset_{})}{\sqrt{\alpha_{s-1}}}\right) + \sqrt{1 - \alpha_{s-1}} \epsilon_{\phi}(x_{s-1}, {s-1}, \emptyset).
\end{aligned}
\end{equation}
Where $\alpha$ represents a set of timesteps with a strictly monotonically increasing noise schedule, hence $\alpha_{s-1} < \alpha_s < \alpha_\mu$. Considering the $\alpha_s < \alpha_\mu$, the term involving $\sqrt{\alpha_\mu}$ should yield a smaller value compared to the term involving $\sqrt{\alpha_s}$, which presents a contradiction to our initial assumptions. Considering the $\epsilon_{\phi}$ terms, due to the inversion process of DDIM, $\epsilon_{\phi}(x_s, s, \emptyset) \approx \epsilon_{\phi}(x_{s-1}, s-1, \emptyset)$ and $\sqrt{1 - \alpha_{s-1}} > \sqrt{1 - \alpha_s}$. This also contradicts the assumption. Therefore, our initial assumption $x_\mu \leq x_s$ must be false. Consequently, we conclude that $x_\mu > x_s$, thus proving the theorem.

\paragraph{ISM is a special case of TSM}
Theoretically, the optimization objectives of ISM and TSM are the same, however, TSM considers optimization steps that are closer together compared to ISM. Specifically, we can regard ISM as a special case of TSM. Between time steps $s$ and $t$, we choose any time step $\mu = \gamma(t-s) + s$, where $\mu = s$ if and only if $\gamma = 0$. Therefore, our hypothesis is confirmed, and ISM is indeed a special case of TSM when $\mu = s$.

\subsection{The abnormal gradient from advanced pipeline}
\label{sec:3.4}
Previous methods have shown that increasing rendering resolution and training batch size can significantly improve visual quality. Although increasing the resolution of rendering can significantly improve the visual quality, most text-to-3D generation methods mainly use guidance based on Stable Diffusion 2.1 and only support $512 \times 512$ resolution. Due to the impact of low resolution, local details are still blurred. Consequently, we experimented with using Stable Diffusion XL as guidance, which supports $1024 \times 1024$ resolutions. The more advanced model Stable Diffusion XL has a different architecture from the previous one, and the VAE of this model is unstable. Although it has a certain impact on NeRF-based methods, it is not serious. However, this instability poses significant challenges for methods that employ 3D Gaussian splatting. In 3D Gaussian splatting, the reliability of operations like copying and deleting point clouds is heavily dependent on gradient stability. If the average positional gradient $g$ of the Gaussian view space exceeds a preset threshold, regions with under- or over-reconstruction of color $c$ and depth $d$ are intensively corrected. SDXL gradients are usually large and unstable, and high average gradient values in this case may cause normal areas to still be densified. An intuitive method is leveraging the gradient clip technical to handle this issue, previous related work \cite{pan2024enhancing} has explored for NeRF-based method, which is not very suitable for 3D Gaussian splitting. Thus, we propose an improved gradient clip method for 3D Gaussian splitting. Specifically, we still use the \cite{pan2024enhancing} method for gradient clipping of color $c$. For depth $d$, we calculate its scaling factor independently for each depth element and perform pixel-by-pixel pruning. The pruning gradient of depth $\boldsymbol{\hat g}_d$ can be expressed as:
\begin{equation}
\boldsymbol{\hat g}_d = \boldsymbol{g}_d \cdot \min\left(\min\left(\frac{s}{|\boldsymbol{g}_d|}, c\right)\right),
\end{equation}
where $\boldsymbol{g}_d$ is the gradient of depth, $s$ is scale of Gaussian and $c$ is the threshold. In this way, we can ensure that the updated direction of the depth gradient remains unchanged and has no effect on the gradient of the normal region.

\begin{figure}
    \centering
    \includegraphics[width=\linewidth]{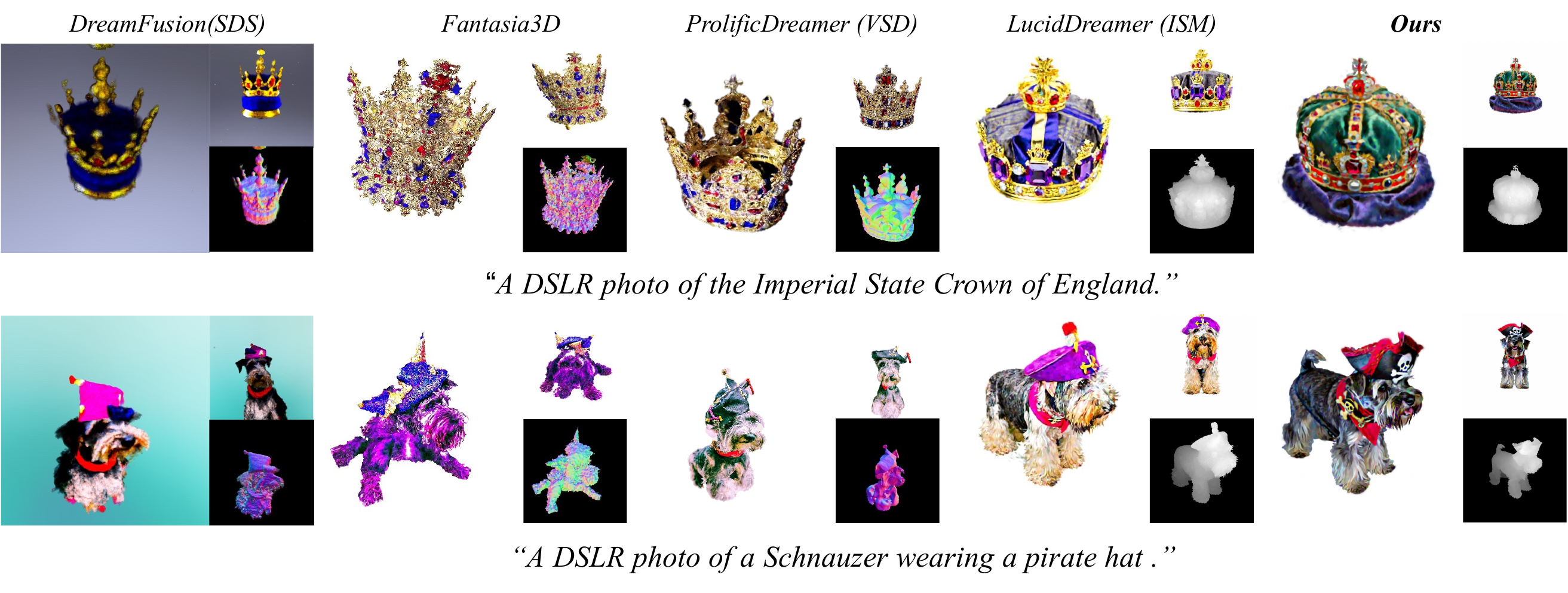}
    \caption{\textbf{Comparison with state-of-the-art baseline methods in text-to-3D generation.} Experimental results show that our method can generate 3D content that is more consistent with input text prompts and has more detailed details. All results of this work are generated on a single A100 GPU. Please zoom in to see more details.}
    \label{fig:2}
\end{figure}
\section{Experiments}
\subsection{Qualitative Results}
\paragraph{Text-to-3D Generation}
We show the generated results of Dreamer XL in \Cref{fig:teaser}. The results show that Dreamer XL is capable of generating high-quality 3D content accurately based on the input text, and it performs exceptionally well in producing realistic and complex appearances, effectively avoiding common issues such as excessive smoothing or oversaturation. For example, it can finely reproduce the texture details of objects like teacups. Moreover, our framework can generate objects that are close to reality and create imaginary ones. This flexibility offers possibilities for various application scenarios.
\paragraph{Comparison with State-of-the-Art Methods} We compare our approach with four state-of-the-art text-to-3D baselines: DreamFusion \cite{poole2022dreamfusion} proposes Score Distillation Sampling (SDS) leveraging a pre-trained 2D text-to-image diffusion model for text-to-3D synthesis; Fantasia3D \cite{chen2023fantasia3d} disentangle geometric and appearance attributes to simulate real-world physical environments;  ProlificDreamer \cite{wang2024prolificdreamer} introduces Variational Score Distillation (VSD), a particle-based variational framework to address issues of oversaturation, oversmoothing, and low diversity; LucidDreamer \cite{liang2023luciddreamer} introduces Interval Score Matching (ISM), utilizing deterministic diffusion trajectories and interval-based score matching to alleviate oversmoothing problems, and employs a 3D Gaussian splatting for 3D representation. The comparison results are shown in \Cref{fig:2}. The results generated by our method are significantly clearer than other baseline results. For example, the crown shows a more precise geometric structure and a more realistic color, and the Schnauzer's hair texture and overall body shape show obvious advantages.  We can observe that our method significantly outperforms existing methods in both visual quality and consistency.

\begin{figure}
    \centering
    \includegraphics[width=\linewidth]{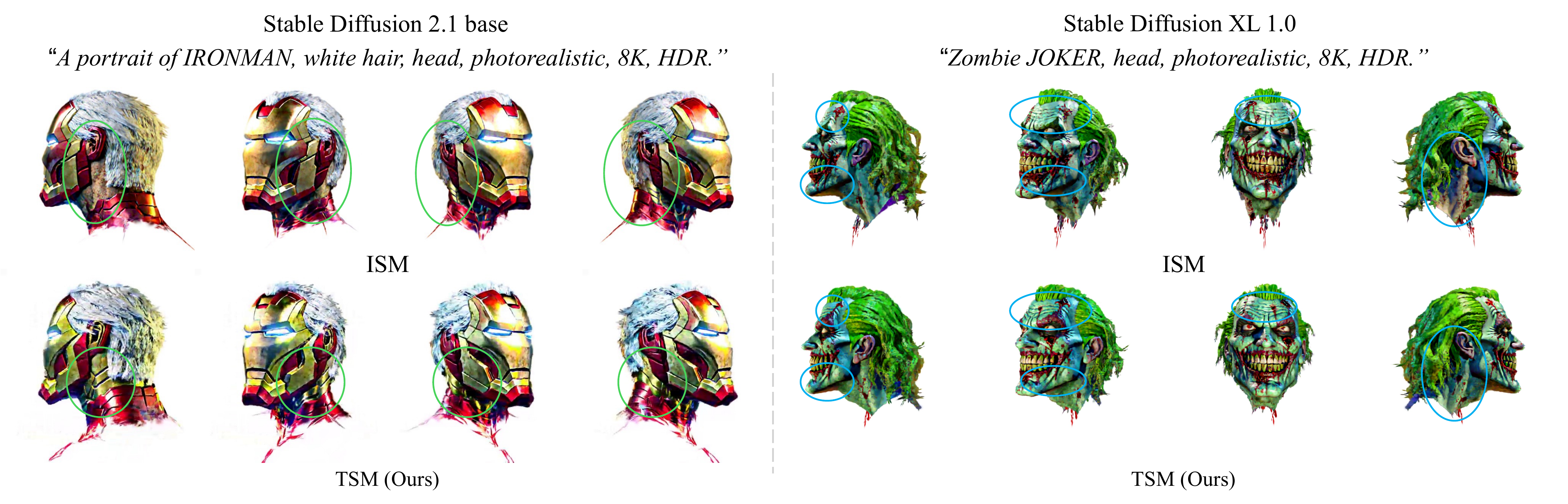}
    \caption{\textbf{Comparison with the generation results of different stable diffusion models.} Compared with ISM, our TSM performs better in the clarity and consistency of local details. Please zoom in to see the circled region for more details.}
    \label{fig:3}
\end{figure}
\paragraph{Comparison with ISM in detail}
As shown in \Cref{fig:3}, we show the generation results of ISM and TSM using the same prompt on different stable diffusion models. In \textit{Iron Man}, it can be seen that the ISM has significant inconsistencies on the left and right sides of the neck, while our TSM maintains consistency in this region. In \textit{Joker}, the ISM has shallower wrinkles on the head compared to our TSM, which is due to the averaging effect caused by error accumulation. Furthermore, ISM also shows significant inconsistency in the neck region.

\begin{table}
\centering
\caption{\textbf{Quantitative evaluation.} We compare with recent text-to-3D conversion methods. CLIP-score is used to measure the alignment between text and 3D content, while A-LPIPS is used to evaluate the degree of artifacts caused by inconsistencies in 3D content.}
\label{table:quan}
\begin{tabular}{lcccc} 
\toprule
\multirow{2}{*}{Methods} & \multicolumn{2}{c}{CLIP-Score $\uparrow$} & \multicolumn{2}{c}{A-LPIPS $\downarrow$} \\
 &  CLIP-L/14 &  OpenCLIP-L/14 &  VGG &   Alex \\
\midrule
DreamFusion & 0.232 & 0.165 & 0.081 & 0.080       \\
Fantasia3D & 0.233 & 0.207 & 0.077 & 0.082      \\
ProlificDreamer & 0.255 & 0.221 & 0.178 & 0.103       \\
LucidDreamer & 0.278 & 0.234 & 0.065 & 0.059       \\
\textbf{Ours} & \textbf{0.297} & \textbf{0.243}& \textbf{0.052}& \textbf{0.041} \\
\bottomrule
\end{tabular}
\end{table} 
\begin{figure}[t]
    \centering
    \includegraphics[width=\linewidth]{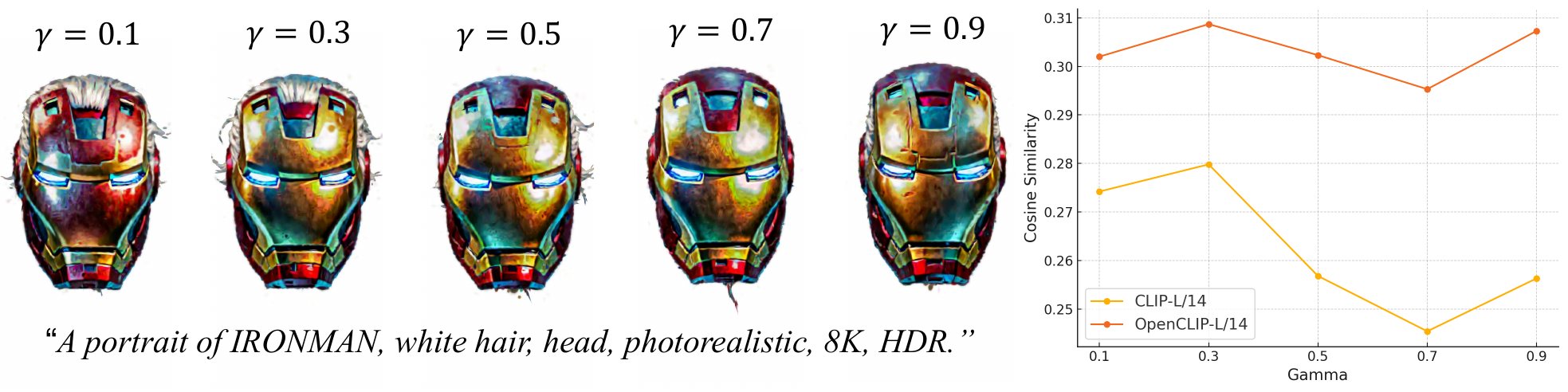}
    \setlength{\abovecaptionskip}{-0.2cm}
    \setlength{\belowcaptionskip}{-0.5cm}
    \caption{\textbf{Ablation} on offset rate. $\gamma=0.3$ achieves optimal visual quality and ensures high consistency between the generated results and the original text.}
    \label{fig:4}
\end{figure}

\subsection{Quantitative Results}
Currently, there are no standardized evaluation metrics specifically dedicated to text-to-3D. This is primarily due to the subjective nature of the task and the presence of multiple dimensions that are difficult to quantify. To maintain consistency with existing text-to-3D evaluation methods, we adopt CLIP-based metrics for quantitative analysis. Specifically, we employ variants of the CLIP model, including OpenCLIP ViT-L/14 and CLIP ViT-L/14, to calculate the average CLIP score between the text and its corresponding 3D render. Furthermore, considering the importance of view consistency, we follow previous work calculating A-LPIPS to determine view consistency, quantifying visual artifacts caused by view inconsistency through calculating the average LPIPS score between adjacent 3D scene images. We adopt A-LPIPS as an alternative metric to quantify view consistency, and present it alongside the CLIP scores in our report.

Consistent with the qualitative results, we compare our method with four state-of-the-art text-to-3D methods. Compared with the current best-performing LucidDreamer, our method improves CLIP-Score (CLIP-L/14) and CLIP-Score (OpenCLIP-L/14) by 6.83\% and 3.85\% respectively. At the same time, our method reduces the performance by 20.00\% and 30.51\% respectively on the A-LPIPS (VGG) and A-LPIPS (Alex) evaluation metrics, showing significant advantages in image authenticity and visual consistency. This improvement is mainly due to our adoption of the more advanced SDXL as a guidance model, which has lower accumulated error. Overall, these results highlight the superior performance of our approach in terms of image quality and text consistency.

\begin{wrapfigure}{r}{0.3\textwidth}
    \centering
    \includegraphics[width=\linewidth]{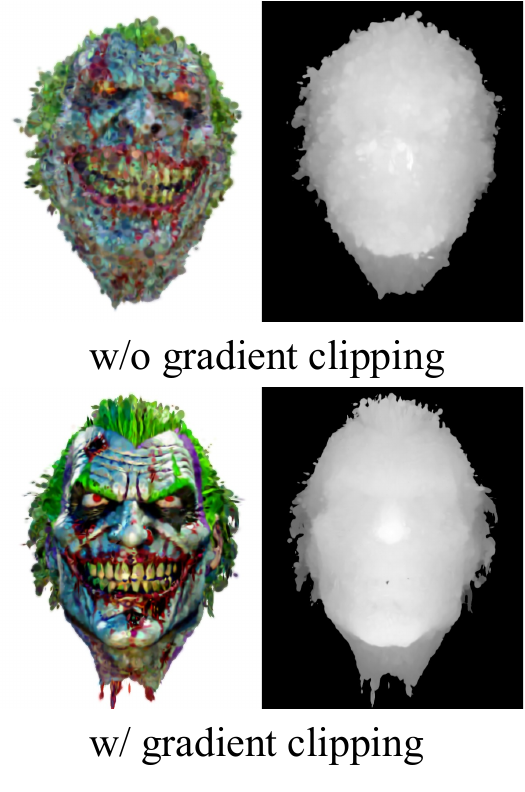}
    \setlength{\abovecaptionskip}{-0.5cm}
    \setlength{\belowcaptionskip}{-0.5cm}
    \caption{\textbf{Ablation} on pixel-by-pixel gradient clipping.}
    \label{fig:5}
\end{wrapfigure}

\subsection{Ablation Study}
\paragraph{Ablation on offset rate $\gamma$}
We investigate the impact of the offset rate $\gamma$ on the generated results (\Cref{fig:4}), and the best results are achieved when $\gamma$ is set to 0.3. If $\gamma$ is set too low, it will result in a loss of color details; if it is set too high, it may destroy the consistency between the generated results and the text. That is, when $\mu$ is too close to $s$, the average effect is too heavy, and its effect is similar to that of ISM. When $\mu$ is too close to $t$, although the cumulative error is reduced, the updated gradient will become very small, easily causing the model to fall into a local optimal state. However, for simple scenes, the results are best when $\mu$ is close to $t$, and the analysis in \Cref{suppab}.

\paragraph{Ablation on pixel-by-pixel gradient clipping} As shown in \Cref{fig:5}, when pixel-by-pixel gradient clipping is not applied, the instability of the gradients causes abnormal splitting and duplication in normal areas, filling the depth map with noise and making it rough and uneven, thus making the entire facial appearance abnormal. However, after applying pixel-by-pixel gradient clipping, it is clearly observed that the depth map becomes smoother, the texture returns to normal, and the normal facial features are displayed. This comparison demonstrates the effectiveness of our method.

\section{Conclusion}
In this work, we investigate the inconsistency problem produced by ISM during the generation of 3D results. To alleviate this problem, we introduce TSM, which leverages dual paths to reduce error accumulation and thereby improve inconsistency. In addition, to simplify the generation of a high-resolution training process, we adopt SDXL as guidance and propose a pixel-by-pixel gradient clipping method to alleviate the abnormal splitting of normal regions in 3D Gaussian splatting caused by SDXL gradient instability. Our experimental results demonstrate that our method can effectively generate high-resolution, high-quality 3D results.




\small

\bibliographystyle{plain}
\bibliography{main}

\begin{thebibliography}{10}

\bibitem{balaji2022ediff}
Yogesh Balaji, Seungjun Nah, Xun Huang, Arash Vahdat, Jiaming Song, Qinsheng Zhang, Karsten Kreis, Miika Aittala, Timo Aila, Samuli Laine, et~al.
\newblock ediff-i: Text-to-image diffusion models with an ensemble of expert denoisers.
\newblock {\em arXiv preprint arXiv:2211.01324}, 2022.

\bibitem{chen2023fantasia3d}
Rui Chen, Yongwei Chen, Ningxin Jiao, and Kui Jia.
\newblock Fantasia3d: Disentangling geometry and appearance for high-quality text-to-3d content creation.
\newblock In {\em Proceedings of the IEEE/CVF International Conference on Computer Vision}, pages 22246--22256, 2023.

\bibitem{dhariwal2021diffusion}
Prafulla Dhariwal and Alexander Nichol.
\newblock Diffusion models beat gans on image synthesis.
\newblock {\em Advances in neural information processing systems}, 34:8780--8794, 2021.

\bibitem{Dynamicduan}
Haoran Duan, Yang Long, Shidong Wang, Haofeng Zhang, Chris~G. Willcocks, and Ling Shao.
\newblock Dynamic unary convolution in transformers.
\newblock {\em IEEE Transactions on Pattern Analysis and Machine Intelligence}, 45(11):12747--12759, 2023.

\bibitem{ho2020denoising}
Jonathan Ho, Ajay Jain, and Pieter Abbeel.
\newblock Denoising diffusion probabilistic models.
\newblock {\em Advances in neural information processing systems}, 33:6840--6851, 2020.

\bibitem{hong2022avatarclip}
Fangzhou Hong, Mingyuan Zhang, Liang Pan, Zhongang Cai, Lei Yang, and Ziwei Liu.
\newblock Avatarclip: Zero-shot text-driven generation and animation of 3d avatars.
\newblock {\em arXiv preprint arXiv:2205.08535}, 2022.

\bibitem{jain2021dreamfields}
Ajay Jain, Ben Mildenhall, Jonathan~T. Barron, Pieter Abbeel, and Ben Poole.
\newblock Zero-shot text-guided object generation with dream fields.
\newblock {\em arXiv}, December 2021.

\bibitem{kerbl3Dgaussians}
Bernhard Kerbl, Georgios Kopanas, Thomas Leimk{\"u}hler, and George Drettakis.
\newblock 3d gaussian splatting for real-time radiance field rendering.
\newblock {\em ACM Transactions on Graphics}, 42(4), July 2023.

\bibitem{kim2023consistency}
Dongjun Kim, Chieh-Hsin Lai, Wei-Hsiang Liao, Naoki Murata, Yuhta Takida, Toshimitsu Uesaka, Yutong He, Yuki Mitsufuji, and Stefano Ermon.
\newblock Consistency trajectory models: Learning probability flow ode trajectory of diffusion.
\newblock {\em arXiv preprint arXiv:2310.02279}, 2023.

\bibitem{liang2023luciddreamer}
Yixun Liang, Xin Yang, Jiantao Lin, Haodong Li, Xiaogang Xu, and Yingcong Chen.
\newblock Luciddreamer: Towards high-fidelity text-to-3d generation via interval score matching.
\newblock {\em arXiv preprint arXiv:2311.11284}, 2023.

\bibitem{lin2023magic3d}
Chen-Hsuan Lin, Jun Gao, Luming Tang, Towaki Takikawa, Xiaohui Zeng, Xun Huang, Karsten Kreis, Sanja Fidler, Ming-Yu Liu, and Tsung-Yi Lin.
\newblock Magic3d: High-resolution text-to-3d content creation.
\newblock In {\em Proceedings of the IEEE/CVF Conference on Computer Vision and Pattern Recognition}, pages 300--309, 2023.

\bibitem{liu2023one}
Minghua Liu, Ruoxi Shi, Linghao Chen, Zhuoyang Zhang, Chao Xu, Xinyue Wei, Hansheng Chen, Chong Zeng, Jiayuan Gu, and Hao Su.
\newblock One-2-3-45++: Fast single image to 3d objects with consistent multi-view generation and 3d diffusion.
\newblock {\em arXiv preprint arXiv:2311.07885}, 2023.

\bibitem{liu2024one}
Minghua Liu, Chao Xu, Haian Jin, Linghao Chen, Mukund Varma~T, Zexiang Xu, and Hao Su.
\newblock One-2-3-45: Any single image to 3d mesh in 45 seconds without per-shape optimization.
\newblock {\em Advances in Neural Information Processing Systems}, 36, 2024.

\bibitem{liu2023zero1to3}
Ruoshi Liu, Rundi Wu, Basile~Van Hoorick, Pavel Tokmakov, Sergey Zakharov, and Carl Vondrick.
\newblock Zero-1-to-3: Zero-shot one image to 3d object, 2023.

\bibitem{luo2023latent}
Simian Luo, Yiqin Tan, Longbo Huang, Jian Li, and Hang Zhao.
\newblock Latent consistency models: Synthesizing high-resolution images with few-step inference.
\newblock {\em arXiv preprint arXiv:2310.04378}, 2023.

\bibitem{ma2023geodream}
Baorui Ma, Haoge Deng, Junsheng Zhou, Yu-Shen Liu, Tiejun Huang, and Xinlong Wang.
\newblock Geodream: Disentangling 2d and geometric priors for high-fidelity and consistent 3d generation.
\newblock {\em arXiv preprint arXiv:2311.17971}, 2023.

\bibitem{metzer2023latent}
Gal Metzer, Elad Richardson, Or~Patashnik, Raja Giryes, and Daniel Cohen-Or.
\newblock Latent-nerf for shape-guided generation of 3d shapes and textures.
\newblock In {\em Proceedings of the IEEE/CVF Conference on Computer Vision and Pattern Recognition}, pages 12663--12673, 2023.

\bibitem{miao2024conrf}
Xingyu Miao, Yang Bai, Haoran Duan, Fan Wan, Yawen Huang, Yang Long, and Yefeng Zheng.
\newblock Conrf: Zero-shot stylization of 3d scenes with conditioned radiation fields, 2024.

\bibitem{michel2022text2mesh}
Oscar Michel, Roi Bar-On, Richard Liu, Sagie Benaim, and Rana Hanocka.
\newblock Text2mesh: Text-driven neural stylization for meshes.
\newblock In {\em Proceedings of the IEEE/CVF Conference on Computer Vision and Pattern Recognition}, pages 13492--13502, 2022.

\bibitem{mildenhall2021nerf}
Ben Mildenhall, Pratul~P Srinivasan, Matthew Tancik, Jonathan~T Barron, Ravi Ramamoorthi, and Ren Ng.
\newblock Nerf: Representing scenes as neural radiance fields for view synthesis.
\newblock {\em Communications of the ACM}, 65(1):99--106, 2021.

\bibitem{nakanishi1999freewalk}
Hideyuki Nakanishi, Chikara Yoshida, Toshikazu Nishimura, and Toru Ishida.
\newblock Freewalk: A 3d virtual space for casual meetings.
\newblock {\em IEEE MultiMedia}, 6(2):20--28, 1999.

\bibitem{pan2024enhancing}
Zijie Pan, Jiachen Lu, Xiatian Zhu, and Li~Zhang.
\newblock Enhancing high-resolution 3d generation through pixel-wise gradient clipping.
\newblock In {\em International Conference on Learning Representations (ICLR)}, 2024.

\bibitem{podell2023sdxl}
Dustin Podell, Zion English, Kyle Lacey, Andreas Blattmann, Tim Dockhorn, Jonas M{\"u}ller, Joe Penna, and Robin Rombach.
\newblock Sdxl: Improving latent diffusion models for high-resolution image synthesis.
\newblock {\em arXiv preprint arXiv:2307.01952}, 2023.

\bibitem{poole2022dreamfusion}
Ben Poole, Ajay Jain, Jonathan~T Barron, and Ben Mildenhall.
\newblock Dreamfusion: Text-to-3d using 2d diffusion.
\newblock {\em arXiv preprint arXiv:2209.14988}, 2022.

\bibitem{qian2023magic123}
Guocheng Qian, Jinjie Mai, Abdullah Hamdi, Jian Ren, Aliaksandr Siarohin, Bing Li, Hsin-Ying Lee, Ivan Skorokhodov, Peter Wonka, Sergey Tulyakov, et~al.
\newblock Magic123: One image to high-quality 3d object generation using both 2d and 3d diffusion priors.
\newblock {\em arXiv preprint arXiv:2306.17843}, 2023.

\bibitem{radford2021learning}
Alec Radford, Jong~Wook Kim, Chris Hallacy, Aditya Ramesh, Gabriel Goh, Sandhini Agarwal, Girish Sastry, Amanda Askell, Pamela Mishkin, Jack Clark, et~al.
\newblock Learning transferable visual models from natural language supervision.
\newblock In {\em International conference on machine learning}, pages 8748--8763. PMLR, 2021.

\bibitem{reisouglu20173d}
Ilknur Reiso{\u{g}}lu, Burcu Topu, Rabia Y{\i}lmaz, T~Karaku{\c{s}}~Y{\i}lmaz, and Yuksel G{\"o}kta{\c{s}}.
\newblock 3d virtual learning environments in education: A meta-review.
\newblock {\em Asia Pacific Education Review}, 18:81--100, 2017.

\bibitem{rombach2022high}
Robin Rombach, Andreas Blattmann, Dominik Lorenz, Patrick Esser, and Bj{\"o}rn Ommer.
\newblock High-resolution image synthesis with latent diffusion models.
\newblock In {\em Proceedings of the IEEE/CVF conference on computer vision and pattern recognition}, pages 10684--10695, 2022.

\bibitem{saharia2022photorealistic}
Chitwan Saharia, William Chan, Saurabh Saxena, Lala Li, Jay Whang, Emily~L Denton, Kamyar Ghasemipour, Raphael Gontijo~Lopes, Burcu Karagol~Ayan, Tim Salimans, et~al.
\newblock Photorealistic text-to-image diffusion models with deep language understanding.
\newblock {\em Advances in neural information processing systems}, 35:36479--36494, 2022.

\bibitem{shi2023zero123++}
Ruoxi Shi, Hansheng Chen, Zhuoyang Zhang, Minghua Liu, Chao Xu, Xinyue Wei, Linghao Chen, Chong Zeng, and Hao Su.
\newblock Zero123++: a single image to consistent multi-view diffusion base model.
\newblock {\em arXiv preprint arXiv:2310.15110}, 2023.

\bibitem{shi2023mvdream}
Yichun Shi, Peng Wang, Jianglong Ye, Mai Long, Kejie Li, and Xiao Yang.
\newblock Mvdream: Multi-view diffusion for 3d generation.
\newblock {\em arXiv preprint arXiv:2308.16512}, 2023.

\bibitem{song2020denoising}
Jiaming Song, Chenlin Meng, and Stefano Ermon.
\newblock Denoising diffusion implicit models.
\newblock {\em arXiv:2010.02502}, October 2020.

\bibitem{song2019generative}
Yang Song and Stefano Ermon.
\newblock Generative modeling by estimating gradients of the data distribution.
\newblock {\em Advances in neural information processing systems}, 32, 2019.

\bibitem{song2020score}
Yang Song, Jascha Sohl-Dickstein, Diederik~P Kingma, Abhishek Kumar, Stefano Ermon, and Ben Poole.
\newblock Score-based generative modeling through stochastic differential equations.
\newblock {\em arXiv preprint arXiv:2011.13456}, 2020.

\bibitem{tang2023dreamgaussian}
Jiaxiang Tang, Jiawei Ren, Hang Zhou, Ziwei Liu, and Gang Zeng.
\newblock Dreamgaussian: Generative gaussian splatting for efficient 3d content creation.
\newblock {\em arXiv preprint arXiv:2309.16653}, 2023.

\bibitem{wang2023score}
Haochen Wang, Xiaodan Du, Jiahao Li, Raymond~A Yeh, and Greg Shakhnarovich.
\newblock Score jacobian chaining: Lifting pretrained 2d diffusion models for 3d generation.
\newblock In {\em Proceedings of the IEEE/CVF Conference on Computer Vision and Pattern Recognition}, pages 12619--12629, 2023.

\bibitem{wang2021neus}
Peng Wang, Lingjie Liu, Yuan Liu, Christian Theobalt, Taku Komura, and Wenping Wang.
\newblock Neus: Learning neural implicit surfaces by volume rendering for multi-view reconstruction.
\newblock {\em arXiv preprint arXiv:2106.10689}, 2021.

\bibitem{wang2024prolificdreamer}
Zhengyi Wang, Cheng Lu, Yikai Wang, Fan Bao, Chongxuan Li, Hang Su, and Jun Zhu.
\newblock Prolificdreamer: High-fidelity and diverse text-to-3d generation with variational score distillation.
\newblock {\em Advances in Neural Information Processing Systems}, 36, 2024.

\bibitem{wodehouse20163d}
Andrew Wodehouse and Mohammed Abba.
\newblock 3d visualisation for online retail: factors in consumer behaviour.
\newblock {\em International Journal of Market Research}, 58(3):451--472, 2016.

\bibitem{wu2024consistent3d}
Zike Wu, Pan Zhou, Xuanyu Yi, Xiaoding Yuan, and Hanwang Zhang.
\newblock Consistent3d: Towards consistent high-fidelity text-to-3d generation with deterministic sampling prior.
\newblock {\em arXiv preprint arXiv:2401.09050}, 2024.

\bibitem{yi2023gaussiandreamer}
Taoran Yi, Jiemin Fang, Junjie Wang, Guanjun Wu, Lingxi Xie, Xiaopeng Zhang, Wenyu Liu, Qi~Tian, and Xinggang Wang.
\newblock Gaussiandreamer: Fast generation from text to 3d gaussians by bridging 2d and 3d diffusion models.
\newblock In {\em CVPR}, 2024.

\bibitem{zhang2023adding}
Lvmin Zhang, Anyi Rao, and Maneesh Agrawala.
\newblock Adding conditional control to text-to-image diffusion models.
\newblock In {\em Proceedings of the IEEE/CVF International Conference on Computer Vision}, pages 3836--3847, 2023.

\bibitem{zhu2023hifa}
Joseph Zhu and Peiye Zhuang.
\newblock Hifa: High-fidelity text-to-3d with advanced diffusion guidance.
\newblock {\em arXiv preprint arXiv:2305.18766}, 2023.

\end{thebibliography}

\clearpage
\appendix

\section{Supplemental material}
\subsection{Implementation details}
During the optimization process, we train the model for 2500 iterations. To optimize the 3D Gaussian model, we set the learning rates for opacity, scaling, and rotation to 0.05, 0.005, and 0.001 respectively. Furthermore, the learning rate of the camera encoder is set to 0.001. During training, RGB images and corresponding depth maps from 3D Gaussians are used for rendering. Gaussian densification and pruning processes are performed between 100 and 1500 iterations. We select the publicly available stable diffusion of text to images as a guidance model and choose the checkpoint of Stable Diffusion XL\footnote{\url{https://huggingface.co/stabilityai/stable-diffusion-xl-base-1.0}}. The guidance scale is 7.5 for all diffusion guidance models.

\subsection{More Qualitative Results}
\begin{figure}[!bth]
    \centering
    \includegraphics[width=\linewidth]{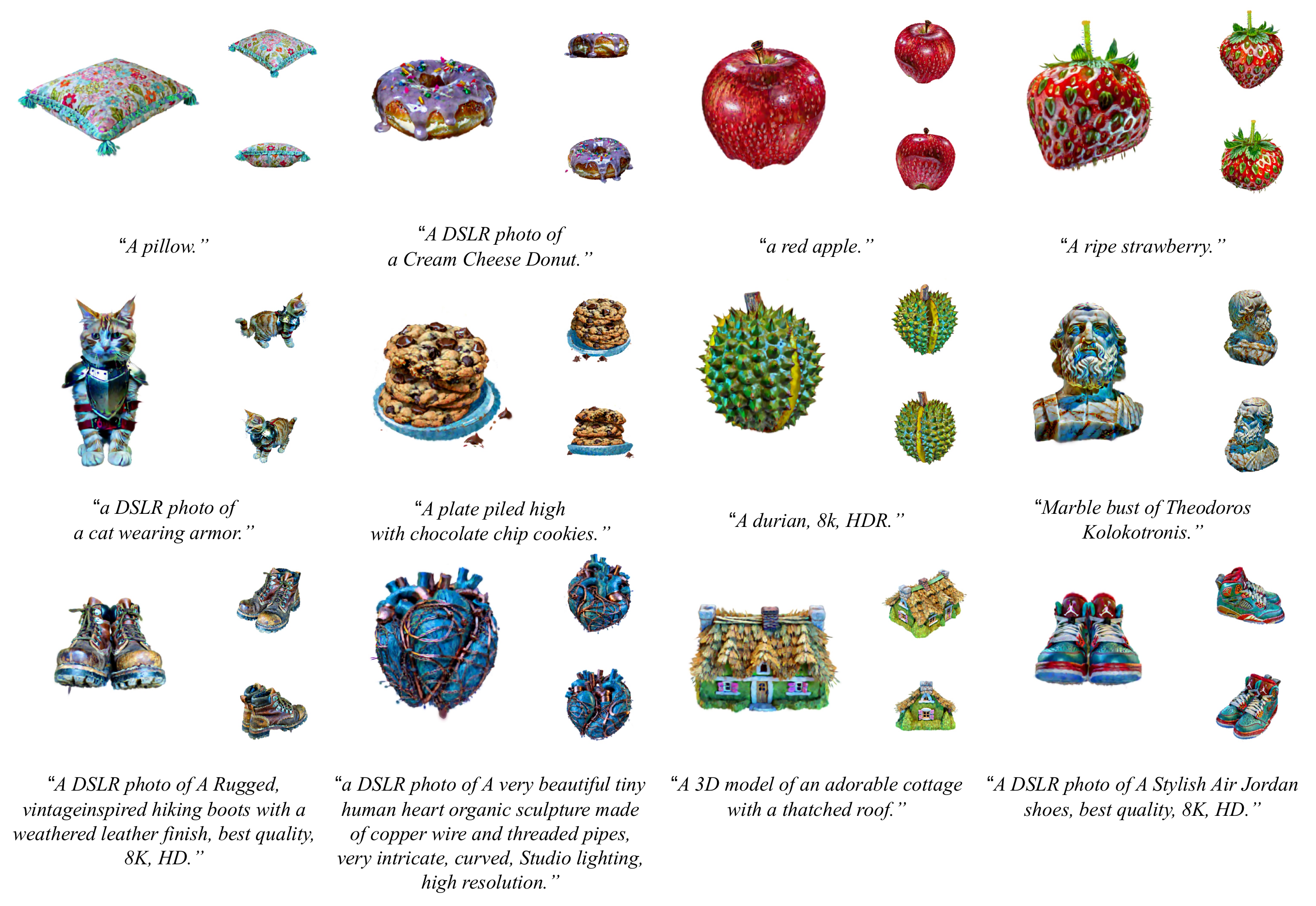}
    \caption{More results generated by our Dreamer XL framework. Please zoom in for details.}
    \label{fig:supp}
\end{figure}

As shown in \Cref{fig:supp}, we present additional generation results. It can be observed that our Dreamer XL is capable of generating 3D models that are visually high-quality, closely approximate reality, and maintain good consistency.

\subsection{More Ablation Study}
\begin{figure}[!bth]
    \centering
    \includegraphics[width=\linewidth]{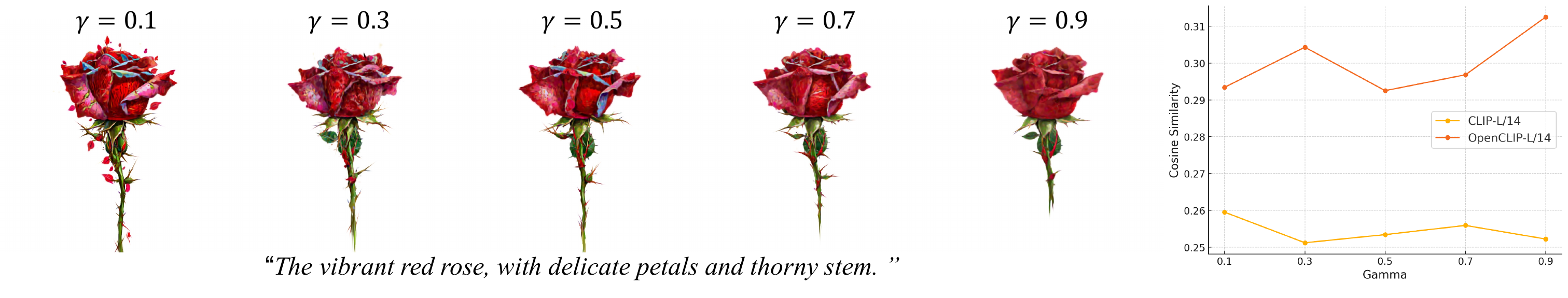}
    \caption{\textbf{More Ablation} on offset rate $\gamma$ for simple scene. }
    \label{fig:supp_ab}
\end{figure}

\begin{figure}[!bth]
    \centering
    \includegraphics[width=\linewidth]{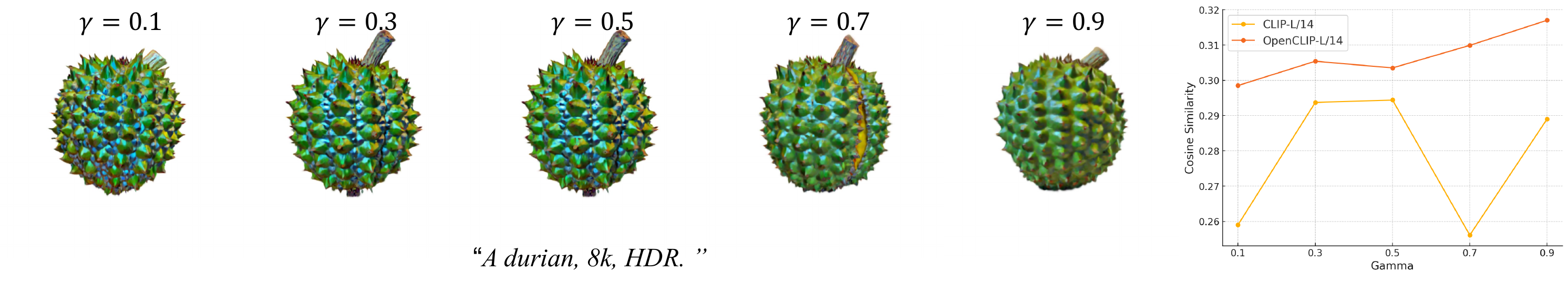}
    \caption{\textbf{More Ablation} on offset rate $\gamma$ for simple scene. }
    \label{fig:supp_ab_1}
\end{figure}

\paragraph{Ablation on offset rate $\gamma$ for simple scene} Here, we provide more ablation on offset rate $\gamma$ for simple scene. As shown in \Cref{fig:supp_ab,fig:supp_ab_1}, if $\gamma$ is set too low, the similarity between text and images decreases, and the images appear unrealistic. Raising $\gamma$ can mitigate the problem of accumulated errors, but it also leads to a loss of some details.

\label{suppab}

\begin{figure}[!bth]
    \centering
    \includegraphics[width=0.6\linewidth]{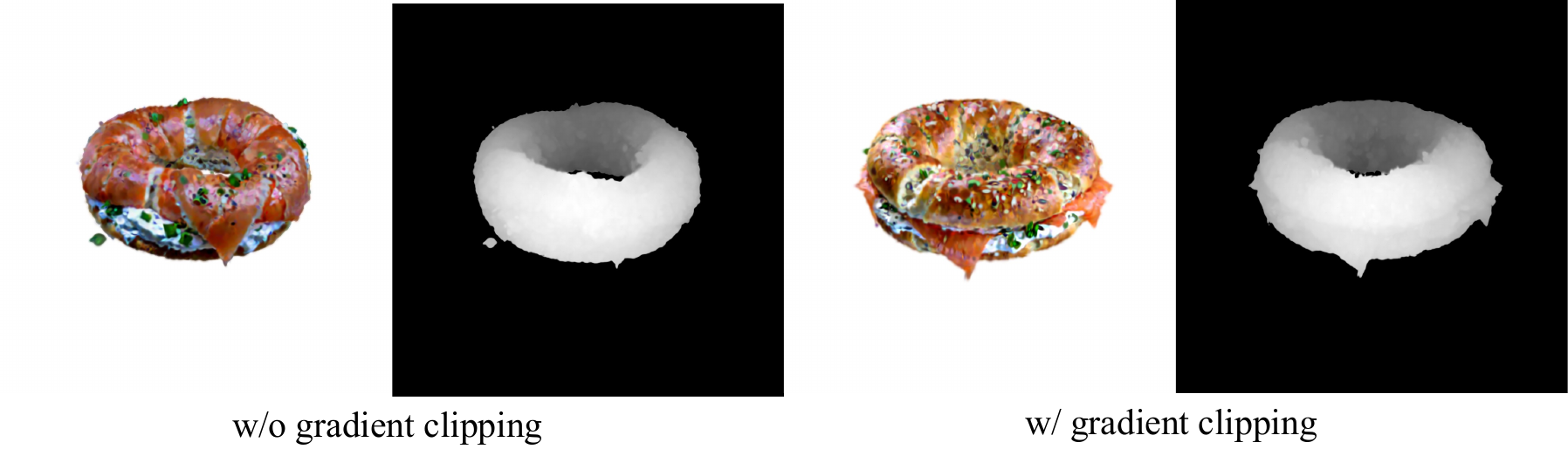}
    \caption{\textbf{More Ablation} on pixel-by-pixel gradient clipping.}
    \label{fig:supp_ab_2}
\end{figure}

\begin{figure}[!bth]
    \centering
    \includegraphics[width=0.6\linewidth]{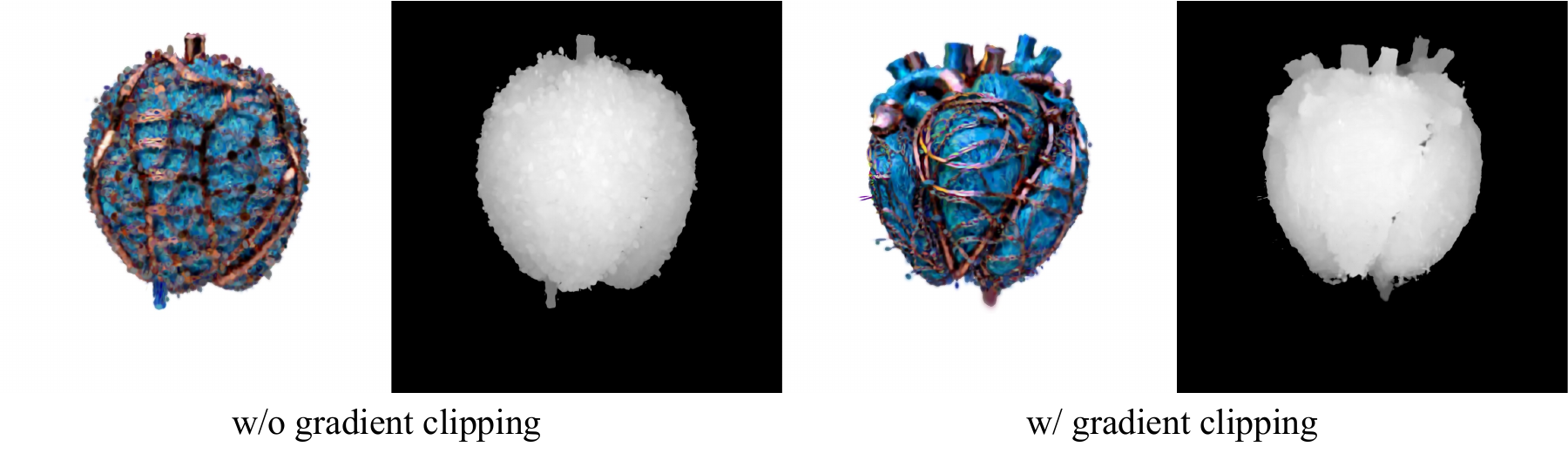}
    \caption{\textbf{More Ablation} on pixel-by-pixel gradient clipping.}
    \label{fig:supp_ab_3}
\end{figure}

\paragraph{Ablation on pixel-by-pixel gradient clipping} Here, we present additional result in \Cref{fig:supp_ab_2,fig:supp_ab_3}. In \Cref{fig:supp_ab_2}, once gradient clipping is not applied, the generated 3D model tends to be darker in color and lacks realism. In \Cref{fig:supp_ab_3}, without gradient clipping, the generated objects appear more blurry, and it is evident from the depth map that there are numerous noise and spikes.

\subsection{Limitations}
While our method can generate high-quality and relatively realistic 3D models, qualitative results show a significant limitation in our approach regarding light handling. Specifically, we have observed anomalous blue reflections in many scenes. Through our experiments, we have identified this problem as primarily caused by our use of SDXL. When SDXL is applied, the blue channel values in rendered images tend to be large, resulting in numerous areas exhibiting abnormal blue hues after normalization. Despite our attempts, including parameter adjustments and different normalization methods, we have yet to find a viable solution. We speculate that this may be attributed to the gradient or training data of SDXL. Additionally, it's worth noting that while our research aims to enhance the quality of generated models, it may inadvertently contribute to the advancement of deepfake technology. 


\end{document}